\documentclass[runningheads]{llncs}
\usepackage{graphicx}

\usepackage{tikz}
\usepackage{comment}
\usepackage{amsmath,amssymb} 
\usepackage{color}
\usepackage{enumitem}
\usepackage{booktabs}
\usepackage{algorithm}
\usepackage{algpseudocode}
\usepackage{setspace}
\usepackage{multirow}
\usepackage{array}
\usepackage{cite}
\usepackage{graphicx}
\usepackage{imakeidx}
\makeindex[name=authors] 

\usepackage[pagebackref=true,breaklinks=true,letterpaper=true,colorlinks,bookmarks=false]{hyperref}
\usepackage[margin=0pt]{caption}
\usepackage[margin=0pt,labelformat=simple]{subcaption}

\renewcommand{\paragraph}[1]{\noindent\textbf{#1}~~}

\usepackage[capitalize]{cleveref}
\crefname{section}{Sec.}{Secs.}
\Crefname{section}{Section}{Sections}
\Crefname{table}{Table}{Tables}
\crefname{table}{Tab.}{Tabs.}

\newcolumntype{P}[1]{>{\centering\arraybackslash}p{#1}}
\newcolumntype{M}[1]{>{\centering\arraybackslash}m{#1}}

\usepackage[accsupp]{axessibility}  

\newcommand{\Tref}[1]{Table~\ref{#1}}
\newcommand{\eref}[1]{Eq.~(\ref{#1})}

\newcommand{\fref}[1]{Fig.~\ref{#1}}

\newcommand{\etal}{et al.~}

\begin{document}

\pagestyle{headings}
\mainmatter
\def\ECCVSubNumber{2166}  

\title{Error Compensation Framework for Flow-Guided Video Inpainting}

\titlerunning{Error Compensation Framework for
Flow-Guided Video Inpainting}
%
\author{Jaeyeon Kang\inst{1} \and
Seoung Wug Oh\inst{2} \and
Seon Joo Kim\inst{1}}

\authorrunning{Kang et al.}
%
\institute{\textsuperscript{1}Yonsei University, \textsuperscript{2}Adobe}
\maketitle

\begin{abstract}
    The key to video inpainting is to use correlation information from as many reference frames as possible.
    Existing flow-based propagation methods split the video synthesis process into multiple steps: flow completion $\rightarrow{}$ pixel propagation $\rightarrow{}$ synthesis.
    However, there is a significant drawback that the errors in each step continue to accumulate and amplify in the next step.
    To this end, we propose an Error Compensation Framework for Flow-guided Video Inpainting (ECFVI), which takes advantage of the flow-based method and offsets its weaknesses.
    We address the weakness with the newly designed flow completion module and the error compensation network that exploits the error guidance map.
    Our approach greatly improves the temporal consistency and the visual quality of the completed videos. 
    Experimental results show the superior performance of our proposed method with the speed up of $\times{6}$, compared to the state-of-the-art methods.
    In addition, we present a new benchmark dataset for evaluation by supplementing the weaknesses of existing test datasets.
\end{abstract}

\section{Introduction}
Video inpainting is the task of filling missing regions in frames with realistic content. 
Its applications cover object removal, watermark/logo/subtitle removal, and even corrupted video restoration. 
It is challenging as the hole regions should be filled with synthesized contents that are unnoticeable with the surrounding background and temporally consistent.

Existing video inpainting methods usually extract useful information from reference frames and merge them to fill the target frame.
The key is to use as many frames as possible since most correlated pixels are somewhere in other frames.
The common pipeline for video inpainting can be categorized into two groups: 1) direct synthesis, 2) flow-based propagation.

\begin{figure}[t]
	\centering

    \begin{subfigure}{\linewidth}
    \centering
    \scriptsize
    \begin{tabular}
        {@{\hskip2pt}p{0.24\linewidth}@{\hskip2pt}p{0.24\linewidth}@{\hskip2pt}p{0.24\linewidth}@{\hskip2pt}p{0.24\linewidth}@{}}
        \centering~Corrupted frame & \centering~STTN\cite{zeng2020learning} &  \centering~ FuseFormer\cite{liu2021fuseformer} & \centering~ECFVI(Ours)
    \end{tabular}
    
    \includegraphics[width=\linewidth]{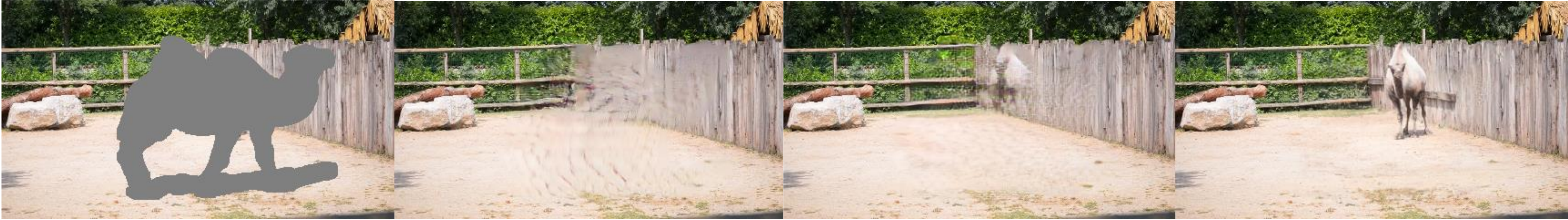}
    \caption{Small temporal window issue in direct synthesis methods}
    \label{fig:teaser1}
    \end{subfigure}
    
    \begin{subfigure}{\linewidth}
    \centering
    \scriptsize
    \begin{tabular}
        {@{\hskip2pt}p{0.24\linewidth}@{\hskip2pt}p{0.24\linewidth}@{\hskip2pt}p{0.24\linewidth}@{\hskip2pt}p{0.24\linewidth}@{}}
        \centering~Corrupted frame & \centering~DFC-Net\cite{xu2019deep} &  \centering~ FGVC\cite{gao2020flow} & \centering~ECFVI(Ours)
    \end{tabular}
    \includegraphics[width=\linewidth]{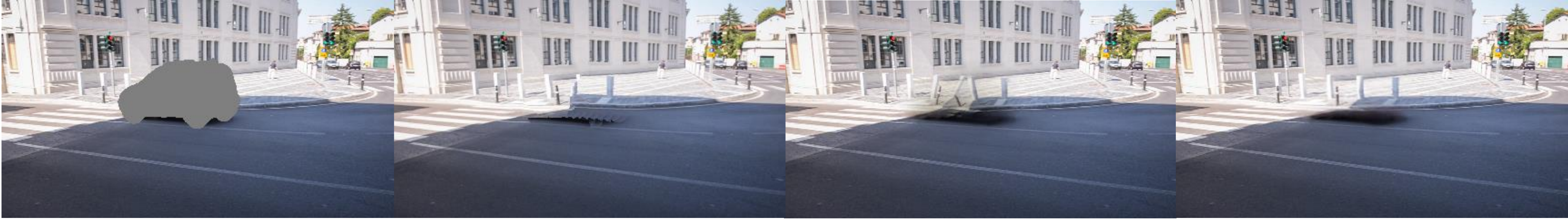}
    \caption{Pixel misalignment and brightness inconsistency issues in flow-based methods}
    \label{fig:teaser2}
    \end{subfigure}
	\caption{Qualitative comparison with other video inpainting methods on object removal scenarios.
	(a) The rear camel is originally shown in all frames, but it disappears in the result of \cite{zeng2020learning, zou2021progressive} due to the small temporal window.
	(b) Wrong pixels are propagated, and the brightness inconsistency issue occurred in \cite{xu2019deep, gao2020flow}.
	}
	\label{fig:teaser}
\end{figure}

In general, direct synthesis methods adopt convolution-based \cite{chang2019free, chang2019learnable, hu2020proposal, zou2021progressive} and attention-based \cite{zeng2020learning, liu2021fuseformer} networks.
They usually take corrupted frames and corresponding pixel-wise masks as input and directly output completed frames.
However, due to the computational complexity, these methods have relatively small temporal windows resulting in only a few reference frames to be used (up to 10 frames).
Serious temporal inconsistencies can occur if some key frames that contain unique pixel values and textures needed to fill the hole are not selected. 
\fref{fig:teaser1} illustrates failure cases of \cite{zeng2020learning, liu2021fuseformer} when an occluded object is not properly included in the small reference frame set. 

Flow-based methods \cite{xu2019deep, gao2020flow} split the  synthesis process as into three steps: flow completion, pixel propagation, and synthesis.
During the flow completion, the corrupted optical flow maps are processed by inpainting the holes within the flow maps.
In the pixel propagation step, pixel values from reference frames are propagated to the holes with the guidance of the inpainted flows.
Finally, in the synthesis step, only the remaining holes are synthesized using an image inpainting method.
Flow-based approaches show better long-term temporal consistency compared to direct synthesis methods as they explicitly propagate pixel values from other frames using flows. 

Nevertheless, there are some drawbacks in flow-based methods. 
\textit{(1) Flow completion.}
In flow inpainting, it is crucial to infer spatio-temporal relationships from other corrupted flows.
However, even adjacent flows can have different values and directions due to inaccurate flow estimation or non-linear motion between frames, resulting in temporally inconsistent flows.
Also, without taking the underlying video contents,  errors from the flow estimator like \textit{From corrupted} in \fref{fig:flow_issue} cannot be handled.
\textit{(2) Pixel propagation.} 
Simply copy-and-pasting propagated pixels causes pixel misalignment and brightness inconsistency.
As the pixel values/textures can be conveyed from distant frames, there are often perspective mismatch such as scale and shape. 
In addition, the brightness inconsistency may arise due to the changes in the lighting condition or camera exposure in the given video sequence.
Models will output visual artifacts without the proper compensation for these errors.
While existing methods attempt to address these issues with checking flow consistency~\cite{xu2019deep, gao2020flow} and poisson blending~\cite{gao2020flow}, visual artifacts still remain as shown in \fref{fig:teaser2}.

To this end, we propose a simple yet effective Error Completion Framework for Flow-guided Video Inpainting (ECFVI) that addresses both drawbacks.
We follow the three steps in flow-based methods (i.e., flow completion, pixel propagation, synthesis) but significantly improve the first two steps by correcting errors, preventing the error accumulation. 
For the flow completion, we make the flow inpainting be aware of RGB values. 
Hallucinating missing flow values based on the underlying video contents can infer temporally coherent flows.
Specifically, we first roughly complete RGB pixels on holes, considering the local temporal relationship.
Then, the locally coherent RGB inpainting results guide the flow inpainting.
Note that the local RGB inpainting here is only used for guiding the flow, and the actual inpainting results are obtained in later steps.  
For the pixel propagation step, we introduce additional compensation network to detect and correct the misalignment and brightness inconsistency errors from the pixel propagation.
The network utilizes the error map outside the hole (called error guidance map) to predict errors inside the hole.   
Such carefully designed components can prevent errors from accumulating at the following stages, which significantly improves the visual quality and temporal consistency of completed videos. 




For quantitative evaluation, many video inpainting methods construct their own datasets by randomly masking regions on a video.
They usually set the unmasked video as the ground-truth and evaluate their models.
However, when the randomly generated mask covers an entire object like in \fref{fig:object_removal}, a model will output object-removed results which are different from the unmasked video.
Evaluation values from this case are not proportional to human perception.
Therefore, we propose a new benchmark dataset by using an object segmentation algorithm \cite{cheng2021stcn} so that the mask only partially covers the object.
The results on our dataset are identical to the unmasked video, which can provide comparative analysis into video inpainting methods.


In summary, the contributions of the paper are as follows:
\begin{itemize} [noitemsep,topsep=0pt,leftmargin=*]

\item We propose a simple yet effective way to compensate for the limitations of the flow-based video inpainting. Specifically, we improve the flow completion with RGB-awareness and propose to compensate errors in pixel propagation. 

\item Our method outperforms previous state-of-the-art video inpainting methods in terms of PSNR/SSIM and visual quality.
Compared with the state-of-the-art in flow-based method FGVC \cite{gao2020flow}, we produce better results while reducing the computation time (×6 faster).
\item We provide a new benchmark dataset for quantitative evaluation of video inpainting, which can be very useful for evaluating future work on this topic by providing a common ground. 
\end{itemize}


\section{Related Work}

\subsection{Direct synthesis methods}
From the success of deep learning, many deep-based video inpainting methods have emerged. 
\cite{wang2019video, kim2019deep, chang2019free} proposed to use 3D encoder-decoder networks for enhancing the efficiency and the temporal consistency.
\cite{zeng2020learning, liu2021fuseformer} exploited attention-based methods, which use transformer modules for matching similar patches on inter-intra frames.
Zou \etal \cite{zou2021progressive} used the optical flow to progressively merge target frames with reference frames to enrich feature representations on temporal relationships. 
Due to the memory constraints, above methods can only refer to a few frames.
On the other hand, Oh \etal \cite{oh2019onion} exploited a non-local pixel matching to have a global temporal window by using the memory network.
Although they refer to all frames, corruption of information occurs because the frames are continuously referenced in an implicit way.
While Lee \etal \cite{lee2019copy} designed a network to take information from long-distance frames with global affine matrices, it cannot handle non-rigid complex motions.

\begin{figure}[t]
	\centering
    \scriptsize
    \begin{tabular}
        {@{}p{0.25\linewidth}@{}p{0.25\linewidth}@{}p{0.25\linewidth}@{}p{0.25\linewidth}@{}}
        \centering~Corrupted frame & \centering~ From original \cite{xu2019deep, gao2020flow} &  \centering~ From corrupted & \centering~Ground-truth flow
    \end{tabular}
    
    \includegraphics[width=\linewidth]{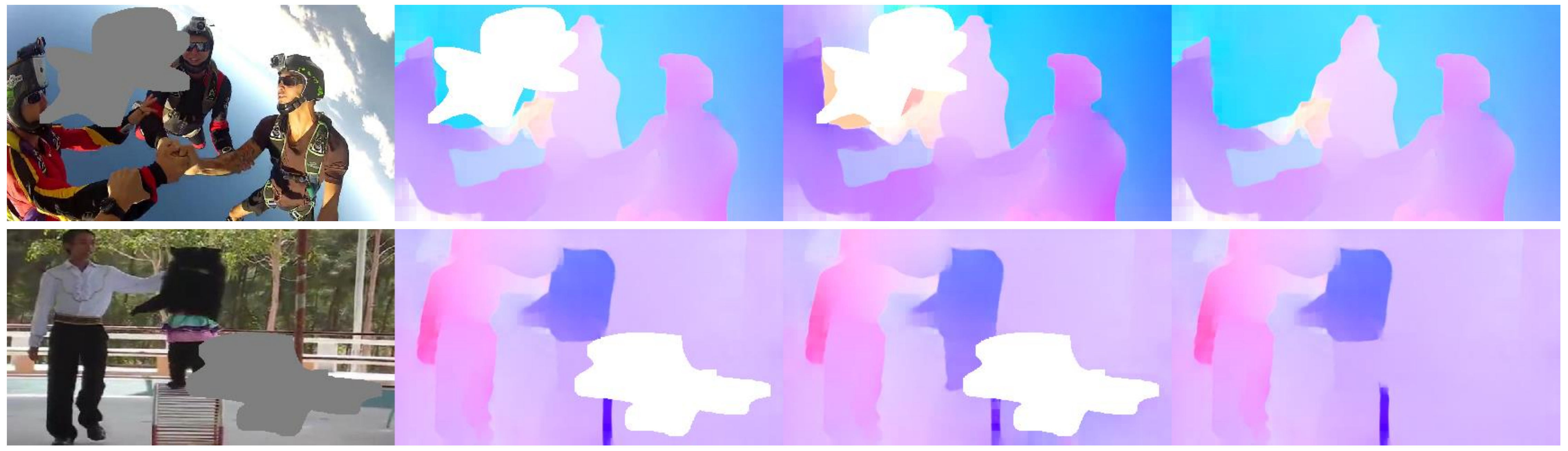}

	\caption{
	This figure shows the corrupted flows with different settings.
	In \cite{xu2019deep,gao2020flow}, they take \textit{original} frames as input to the flow estimator and remove the values in the hole.
	However, if the original frames are unavailable, errors can occur when the corrupted frames are used as input.
	The flow values near the hole are different from the ground-truth flow.
	}
	\label{fig:flow_issue}
\end{figure}

\subsection{Flow-based methods} \label{sssec:Flow-based methods}
Huang \etal \cite{huang2016temporally} adopted spatial patch matching and flow-based method but suffered from mismatching issues caused by corrupted flows.
Instead, \cite{xu2018youtube, gao2020flow} first inpainted the corrupted flows and propagated pixels along their flow trajectories.
Despite the success, they still have some limitations to produce better results:
First, to estimate the corrupted flows, they entered \textit{original} frames to the flow estimator and removed the values with masks.
This procedure works well in object removal scenarios where the original frames exist.
The problem is when the original frames are not available as in video restoration scenarios.
If the flow estimator takes the corrupted frames as input, the masks can interfere with estimating the motion of objects in the video (\fref{fig:flow_issue}).
The errors in corrupted flows will affect the flow completion stage,  resulting in incorrect flow values. 
Second, it is difficult to use spatio-temporal information between the corrupted flows for flow completion as mentioned in previous section.


Wrong estimation from the flow completion would lead to the pixel misalignment issue.
Both methods checked the flow consistency and only propagated trustworthy pixels to deal with the issue, but the consistency is not well preserved as the completed flow itself is incorrect. 
Our framework focuses on offsetting these limitations of flow-based methods.

\section{Method}
Let $X=[x_1, x_2, ..., x_T]$ be a set of corrupted video frames of length $T$.
$M=[m_1, m_2, ..., m_T]$ denotes the corresponding frame-wise masks, where the spatial resolution is the same as $X$. 
For each mask, value $0$ indicates valid regions, and $1$ indicates hole (missing) regions.
Video inpainting aims to predict the original or object removed video $\hat{Y}=[\hat{y}_1, \hat{y}_2, ..., \hat{y}_T]$ given $X$ and $M$ as inputs.
We call the frame to be inpainted as the target frame and other frames as reference frames.

\begin{figure}[t]
    \centering
    \includegraphics[width=\linewidth]{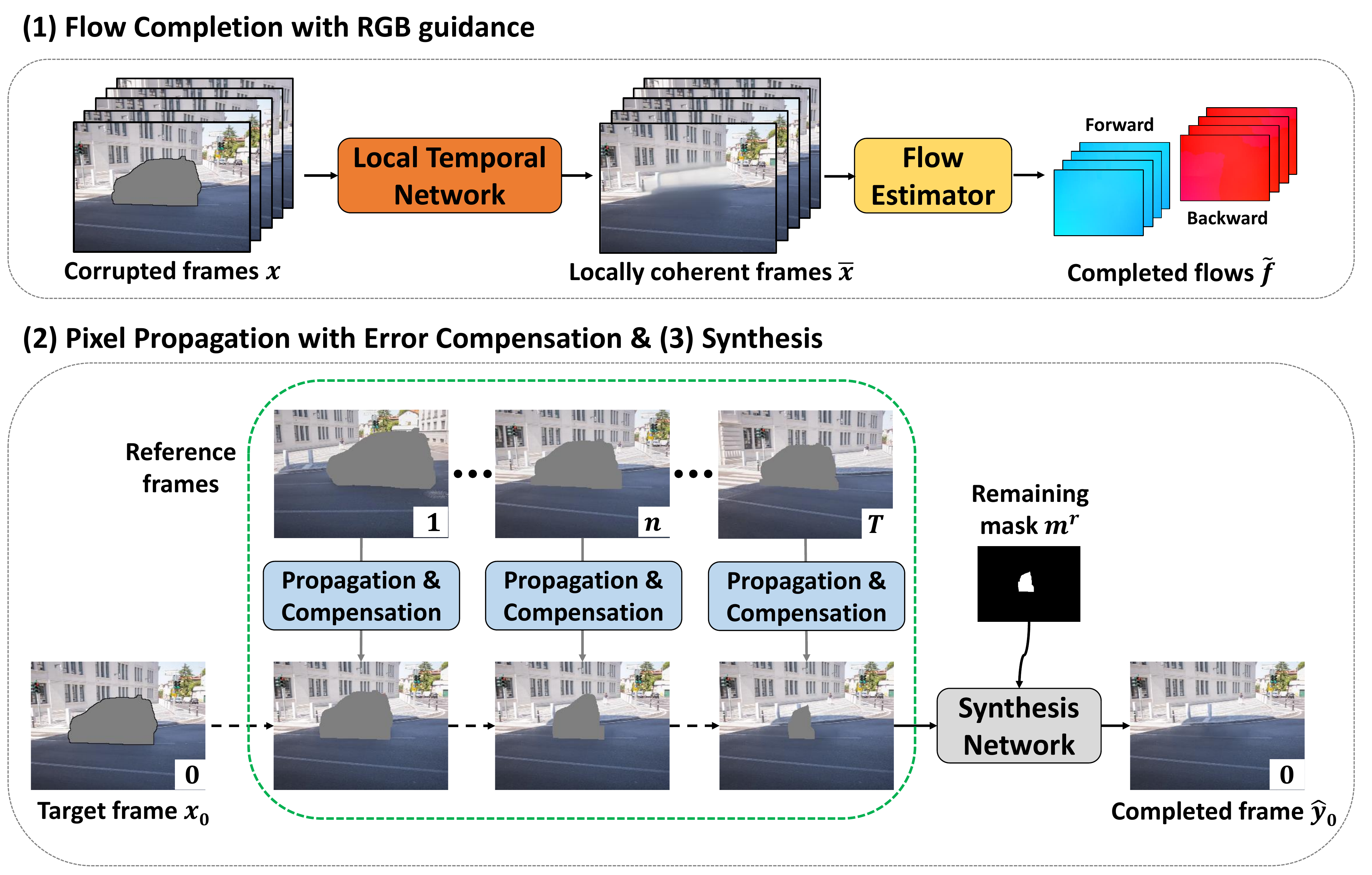}
	\caption{Overview of our Error Compensation Framework for Flow-guided Video Inpainting (ECFVI). 
	At first, we estimate all completed flows between adjacent frames using our flow completion module.
	With the guidance of the completed flows, we iteratively propagate and compensate pixels from reference frames to fill target holes.
	One iteration of the process is shown in \fref{fig:FillandRefine}.
	The remaining holes are further synthesized with the existing video inpainting method.
	}
	\label{fig:overview}
\end{figure}

\subsection{Overview}
%
We illustrate our video inpainting framework in \fref{fig:overview}.
The overall procedure can be summarized as follows.
\textbf{(1) Flow completion with RGB guidance: } 
Given the corrupted frames, we first generate locally coherent frames.
Then, we estimate complete flows between adjacent frames using our flow estimator (Sec. \ref{sssec:Flow Completion}).
Before the propagation, we estimate bi-directional completed flows between all frames.
\textbf{(2) Pixel Propagation with Error Compensation: }
With the guidance of completed flows, we propagate valid pixels from reference frames to target holes. 
In this process, we prevent the errors from propagating to the following stages using our error compensation network (Sec \ref{sssec:Propagation and Compensation}). 
In the order close to the target frame, we iterate propagation and compensation procedures until the hole regions are entirely removed or all frames are referenced.
\textbf{(3) Synthesis: } 
If there are remaining holes to fill, we synthesize using an existing video inpainting method. 
This case usually occurs due to occlusion, where some pixels cannot be found in any other frames.
For the synthesis, we used FuseFormer \cite{liu2021fuseformer} with the weights from their website.


\subsection{Flow completion with RGB guidance} \label{sssec:Flow Completion}
Previous methods \cite{xu2019deep, gao2020flow} first estimate corrupted flows with the flow estimator \cite{ilg2017flownet, teed2020raft} and complete flow values with their methods.
However, without considering the original video contents, there is no ability to handle the errors from the corrupted flows.
Therefore, we design our flow completion module to be aware of RGB values and directly output completed flows from the flow estimator. 


A simple approach would be to have the flow estimator take the corrupted frames and complete the flow values.
However, since the flow estimator \cite{ilg2017flownet, teed2020raft} iteratively takes two frames to estimate entire flows, completed flows are created without considering spatio-temporal information of other flows.
Also, it is more challenging to run flow estimation and completion simultaneously.
Therefore, we roughly complete RGB pixels through local neighboring frames (larger than two frames) and pass the results to our flow estimator.
This can further exploit temporal information for estimating completed flows.
Here, our flow estimator takes locally coherent frames $\bar{x}$ from a local temporal network $LTN$.
Note that the intermediate frames $\bar{x}$ are only used for guiding the flow completion.


We design the local temporal network $LTN$ to consist of an encoder, multiple spatio-temporal transformer layers \cite{zeng2020learning, liu2021fuseformer} and a decoder.
We first extract features of each input from the encoder and exploit transformer layers to merge the information from the reference in the deep encoding space.
The decoder takes the output of the transformer layers to reconstruct locally coherent frames $\bar{x}$ as follows:
\begin{equation}
    \begin{aligned}
        \bar{x}_{i} = LTN(x_{i-N:i+N},m_{i-N:i+N}),
    \end{aligned}
    \label{eq:Coarse network}
\end{equation}
where $N$ is the temporal radius and $N=5$ is used in our experiment.
Then, we estimate a completed flow between adjacent frames as follows:
\begin{equation}
    \begin{aligned}
        \tilde{f}_{t\rightarrow{t+1}} = F(\bar{x}_{t}, \bar{x}_{t+1}, m_{t}, m_{t+1}),
    \end{aligned}
    \label{eq:Coarse network}
\end{equation}
where $F$ is our flow estimator. 
Backward flow $\tilde{f}_{t\rightarrow{t-1}}$ can be estimated in the same manner.
For classifying the regions where the original and coarsely completed content exist, we use masks $m$ as  additional input.
The flow estimator is initialized with pretrained weights from RAFT \cite{teed2020raft} except for the first layer to take the masks as additional input.
We jointly train our local temporal network and flow estimator as follows:
\begin{equation}
    \begin{aligned}
        \mathcal{L} =  \mathcal{L}_{rec} +  \lambda_{flow}\mathcal{L}_{flow},
    \end{aligned}
    \label{eq:loss of flow completion}
\end{equation}
where $\mathcal{L}_{rec}$ is the reconstruction L1 loss defined as $||\bar{x}-{y}||_{1}$. 
$\lambda_{flow}$ is a coefficient for the loss terms and we set it to $2$ in, our experiment.
For the flow loss $\mathcal{L}_{flow}$, inspired by \cite{shrivastava2016training}, we employ hard example mining mechanism to weigh more on the difficult areas as follows:
\begin{equation}
    \begin{aligned}
        \mathcal{L}_{flow}  =  || {h}_{t} \odot{} (\tilde{f}_{t \rightarrow{t+1}} - f_{t \rightarrow{t+1}})||_{1},
    \end{aligned}
    \label{eq:loss of flow spec}
\end{equation}
where $\odot{}$ is element-wise multiplication and ground-truth $f_{t \rightarrow{t+1}}$ is calculated from original frames $Y$.
We set the hard mining weight ${h}_{t}$ with $ (1 +|\bar{x}_{t} - y_{t}|)^2$.
It encourages the model to implicitly deal with errors from the local temporal network. 
More details of our network are shown in the supplementary material.

This design choice has advantages in two aspects.
First, it alleviates the problems of corrupted flows mentioned in Sec.\ref{sssec:Flow-based methods}.
Second, with the strong initialization of pretrained weights and no need to inpaint flow values from scratch, it can be trained much easier.
Although there still could be errors from the two networks, we can further compensate it in the next stage.

\begin{figure*}[t]
    \centering	
    
    \centering

    \includegraphics[width=\linewidth]{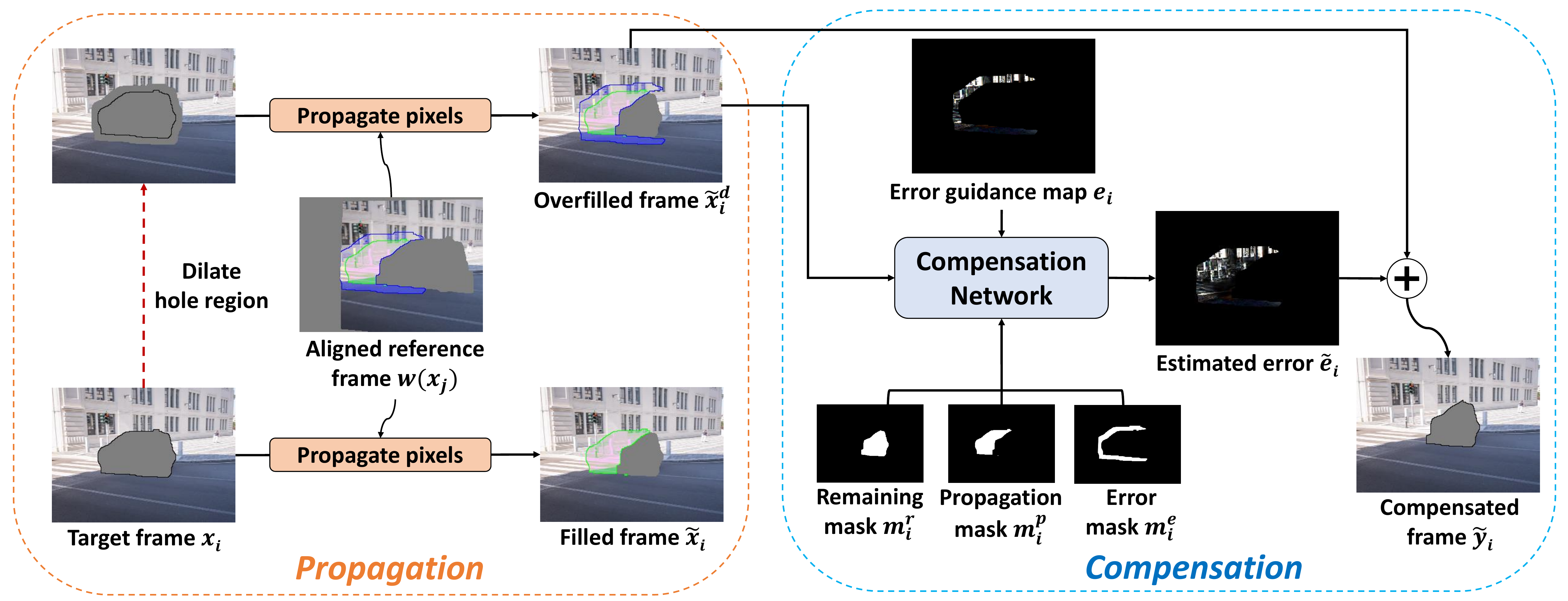}
    

	\caption{
	For simplicity, we show the propagation and compensation stages from one reference frame $x_j$ to the target frame $x_i$.
	We dilate the original hole region to get the overfilled frame $\tilde{x}^{d}_i$.
	Pixels corresponding to the blue region are further propagated from the aligned reference frame.
	The filled frame $\tilde{x}_i$ and overfilled frame $\tilde{x}^{d}_i$ are used to estimate the error guidance map $e_i$.
	Note that the green and blue regions correspond to propagation and error mask respectively.
	}
	\label{fig:FillandRefine}
\end{figure*}

\subsection{Pixel Propagation with Error Compensation} \label{sssec:Propagation and Compensation}
The next step, pixel propagation with error compensation, is illustrated in \fref{fig:FillandRefine}.
With the guidance of the completed flows, we iterate propagation and compensation steps so that the errors do not accumulate or amplify in the next iteration.

\paragraph{Propagation.} 
Let $x_i$ and $x_j$ be the target and the reference frame respectively.
With the completed flow $\tilde{f}_{i\rightarrow{j}}$, only the valid pixels in reference frame $x_j$ are forwarded to target holes by using backward warping function $w$. 
\begin{equation}
    \begin{aligned}
        {m}^{p}_{i} = m_i \odot{}(1-w(m_j,\tilde{f}_{i\rightarrow{j}})), & \\
        \tilde{x}_i = x_i + {m}^{p}_{i}\odot{}w(x_j, \tilde{f}_{i\rightarrow{j}}),
    \end{aligned}
    \label{eq:propagate pixel}
\end{equation}
where $\tilde{x}_i$ is the filled target frame.
To find the regions where the matched and valid regions are in the reference frame, we warp the reference mask $m_j$ and get valid regions $1-w(m_j,\tilde{f}_{i\rightarrow{j}})$ in the warped reference frame.
$m^p_i$ denotes a propagation mask where the propagated pixels exist, shown by the green region in \fref{fig:FillandRefine}.
This mask is used at the compensation stage to know where to compensate for errors.
After the propagation, there may be a remaining mask $m^{r}_{i}$, where $m^{r}_{i} = m_i - {m}^{p}_{i}$.
The remaining mask is further used for the next propagation or input for the synthesis network.

\begin{figure}[t]
	\centering
    \includegraphics[width=1.0\linewidth, scale=0.8]{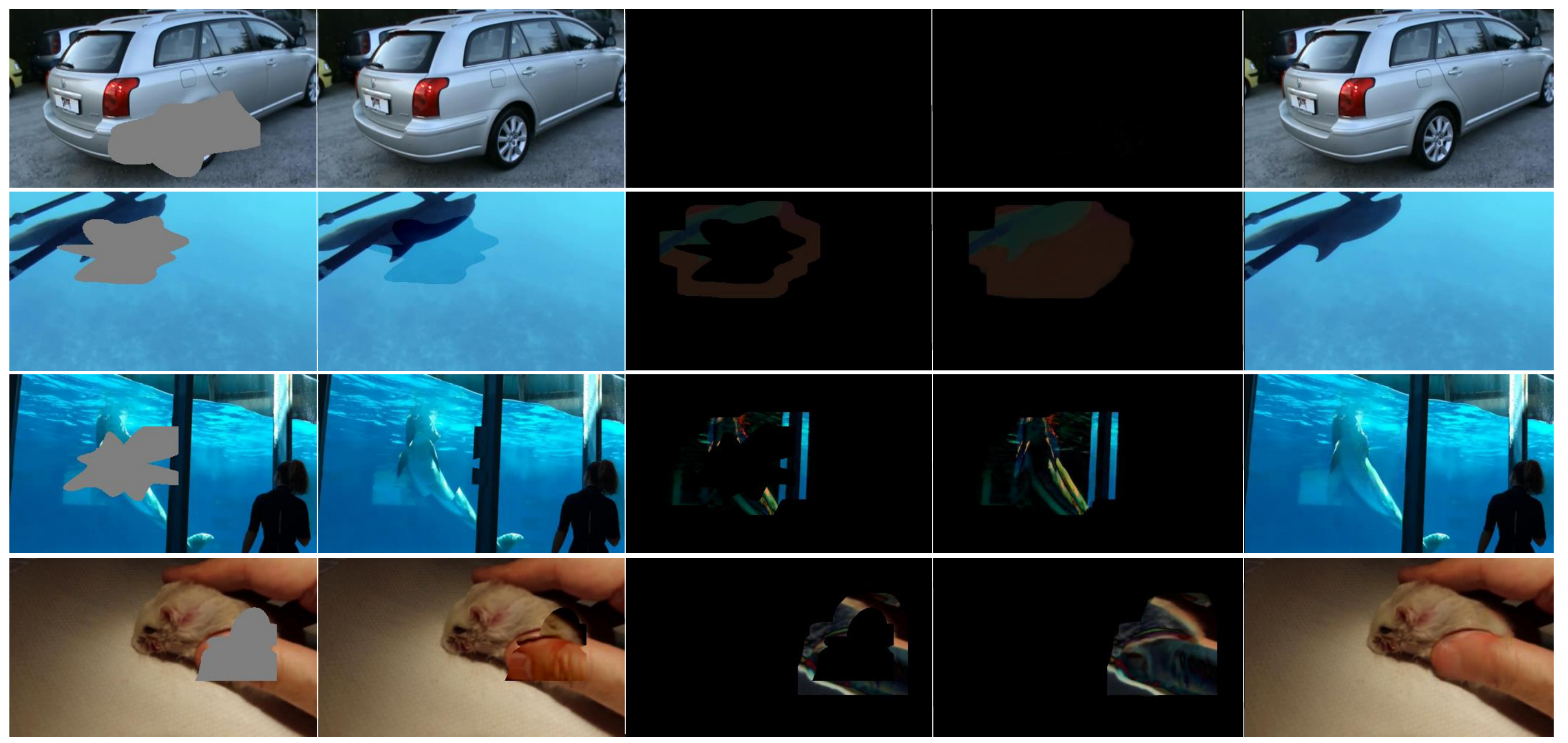}
    \scriptsize
    \begin{tabular}
        {@{\hskip2pt}p{0.19\linewidth}@{\hskip2pt}p{0.19\linewidth}@{\hskip2pt}p{0.185\linewidth}@{\hskip2pt}p{0.205\linewidth}@{\hskip2pt}p{0.175\linewidth}@{}}
        \centering~Target frame  \( x_i\) & 
        \centering~Filled frame  \(\tilde{x}_i\) &
        \centering~Error guidance \\ map  \(e_i\)  &
        \centering~Estimated error  \(\tilde{e}_i\) &
        \centering~Compensated \\ frame \(\tilde{y}_i\)
    \end{tabular}
	\caption{
	Problems from the propagation stage and corresponding error maps.
	From top to bottom:  no issue, brightness inconsistency, pixel misalignment, and combined issues.
	The estimated errors and compensated frames are computed using our method.
	}
	\label{fig:Error_map}
\end{figure}

\paragraph{Compensation.}
As shown in \fref{fig:Error_map}, we observe that directly propagating pixels from the reference frame may cause misalignment or brightness inconsistency issues.
To remedy these issues, one solution is taking filled frame $\tilde{x}_i$ and corresponding mask $m^p_i$ as input and processing them through a GAN framework.
However, it is difficult to train such a network because it cannot easily detect what types of problem occurred.

We approach this problem by introducing the error guidance map $e_i$ as input to our network.
To know what is wrong with the propagation, we need valid (ground-truth) values.
Since only regions outside the holes in the given corrupted frame have valid values, we propagate more pixels by dilating the original hole regions like in \fref{fig:FillandRefine}.
Then, we calculate the errors on enlarged parts, which is the error guidance map $e_i$.
This process assumes that the propagated pixels on the enlarged and the propagated regions have similar error tendencies.
We set the dilation factor as 17 pixels in our experiment. 

 
We set dilated mask as $m^d_i$ and overfilled frame as $\tilde{x}^d_i$.
The error guidance map $e_i$ and the corresponding error regions (mask) $m^e_i$ are computed as follows:
\begin{equation}
    \begin{gathered}
        m^e_i = (m^d_i -m_i) \odot{} (1- w(m_j,\tilde{f}_{i\rightarrow{j}})),  \\
        e_i = m^e_i\odot{}(\tilde{x}^d_i - \tilde{x}_i).
    \end{gathered}
    \label{eq:error map}
\end{equation}
In \fref{fig:Error_map}, this error guidance map intuitively shows the problems of pixel propagation.
If it has no issue, it will be black (zero).
On the other hand, if it has a pixel misalignment issue, it will result in the edges being visible in the error map.
This additional information helps our network be more aware of the problems from the propagation stage.

We design our error compensation network with a similar structure as our local temporal network $LTN$, but with different input and output settings:
\begin{equation}
    \begin{aligned}
        \tilde{e}_i = ECN(\tilde{x}^d_{i-N:i+N}, {e}_{i-N:i+N}, m^{[e,p,r]}_{i-N:i+N}),
    \end{aligned}
    \label{eq:refinement network2}
\end{equation}
where $ECN$ is our error compensation network and $m^{[e,p,r]}$ denotes error, propagation, remaining mask respectively.
For simplicity, we will omit the index from now on.
$\tilde{x}^d, {e}, m^e$ are concatenated in the channel dimension and forwarded to our encoder, and each of the two masks $m^p, m^r$ are used at the transformer layers.
Note that the overfilled frames $\tilde{x}^d$ are used instead of filled frames $\tilde{x}$ as input.
It makes our network to learn the relationship between overfilled frame $\tilde{x}^d$ and the error guidance maps $e$ on the error regions $m^e$.
Since the propagated regions have some structure information, based on the relationship,
our network can estimate error values on the regions $m^p$ with accuracy.
A final compensated frame is $\tilde{y} = \tilde{x}^d + \tilde{e}$.
The results and corresponding errors are shown in \fref{fig:Error_map}.

The loss to train our error compensation network consists of reconstruction loss $\mathcal{L}_{rec}$ and adversarial loss $\mathcal{L}_{adv}$:
\begin{equation}
    \begin{aligned}
        \mathcal{L}  =  \mathcal{L}_{rec} + \lambda_{adv}\mathcal{L}_{adv},
    \end{aligned}
    \label{eq:loss of refinement network}
\end{equation}
where $\lambda_{adv}$ is a coefficient for the loss terms and we set $0.01$ in our experiment.
$\mathcal{L}_{rec}$ is a L1 loss between the ground-truth frame ${y}_i$ and compensated output $\tilde{y}_i$.
For $\mathcal{L}_{adv}$, we use the model from Temporal PatchGAN (T-PatchGAN) \cite{chang2019free}. 
Further details on our network are included in the supplementary material.

\section{Experiments}

\subsection{Training Details}
Our whole networks are trained on Youtube-VOS \cite{xu2018youtube} dataset. 
It consists of 4,453 videos, which are split into 3,471/474/508 for training, validation and testing respectively. 
We randomly select $256\times{}256$ frames and free-form masks from \cite{zeng2020learning} as input.
To train our compensation network, we freeze the weights of flow completion. 
And for handling the brightness inconsistency issue, we further modify the brightness, hue and saturation of reference frames.
Adam optimizer \cite{kingma2014adam} is used where the learning rate starts with 1e-4 and is divided by 2 every 40,000 iterations.
For each module, the total iteration is 120,000 with mini-batch size 4, and it takes about 2 days using two NVIDIA RTX 2080 TI GPUs.

\begin{figure}[t]

    \begin{subfigure}{\linewidth}
    \centering
    \small
    \begin{tabular}
        {@{}p{0.3\linewidth}@{}p{0.3\linewidth}@{}p{0.3\linewidth}}
          \centering~Original frame & \centering~Corrupted frame  & \centering~ECFVI(Ours)
    \end{tabular}
    
    \includegraphics[width=0.9\linewidth]{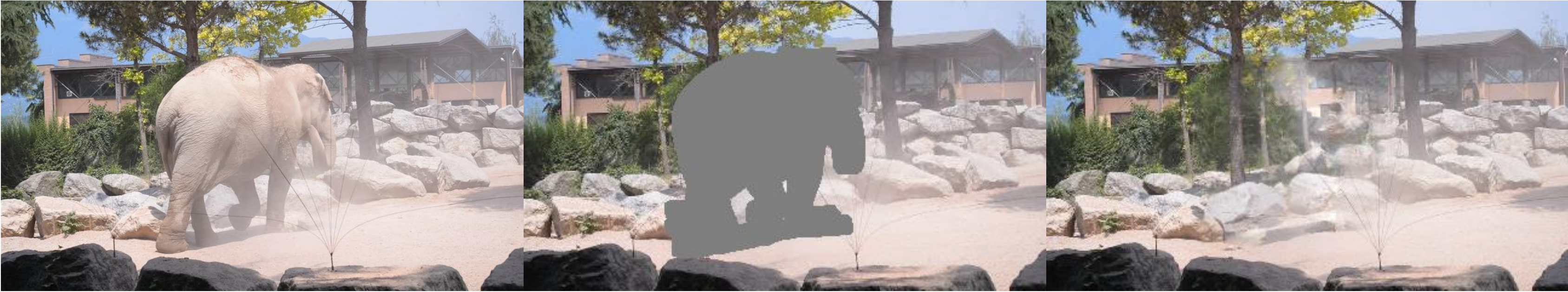}
    \caption{Object removal scenario}
    \label{fig:object_removal}
    \end{subfigure}
    
    \begin{subfigure}{\linewidth}
    \centering
    \includegraphics[width=0.9\linewidth]{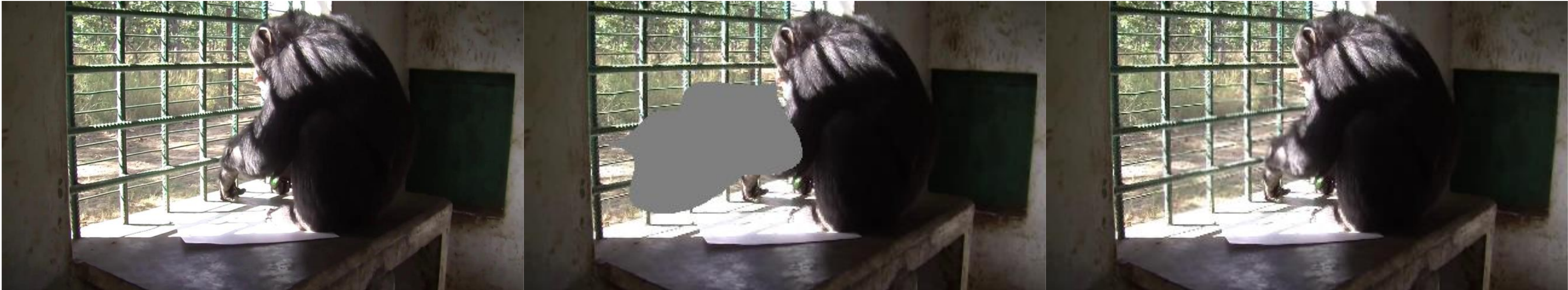}
    \caption{Video restoration scenario}
    \label{fig:video_restoration}
    \end{subfigure}
    
	\caption{
	Two common scenarios for video inpainting: object removal and video restoration.
	The gray region in the corrupted frame denotes a mask.
	In the object removal scenario, the result and the original frame are different.
	Only the video restoration scenario should be used for qualitative evaluation.
	}
	\label{fig:two_scenarios}
\end{figure}

\subsection{Youtube-VI Dataset}
Most previous works created their own random masks and measured the performance on Youtube-VOS \cite{xu2018youtube} and DAVIS \cite{caelles20182018} datasets.
However, when a randomly generated mask coincidentally covers an entire object, a model will create the object-removed result like in \fref{fig:object_removal}.
The result is reasonable, but it is far from the original frame.
Only the case in \fref{fig:video_restoration} should be used for quantitative evaluation.
Also, if there is no motion in the mask and its vicinity, it is not suitable for video inpainting, which cannot trace any pixels from other frames.

\begin{figure}[t]
	\centering
	
	\scriptsize
    \begin{tabular}
        {@{}p{0.25\linewidth}@{}p{0.25\linewidth}@{}p{0.25\linewidth}@{}p{0.25\linewidth}}
        \centering~(a) Circle & \centering~(b) Curve &  \centering~ (c) Free-form & \centering~ (d) Stationary
    \end{tabular}
		\includegraphics[width=1\linewidth]{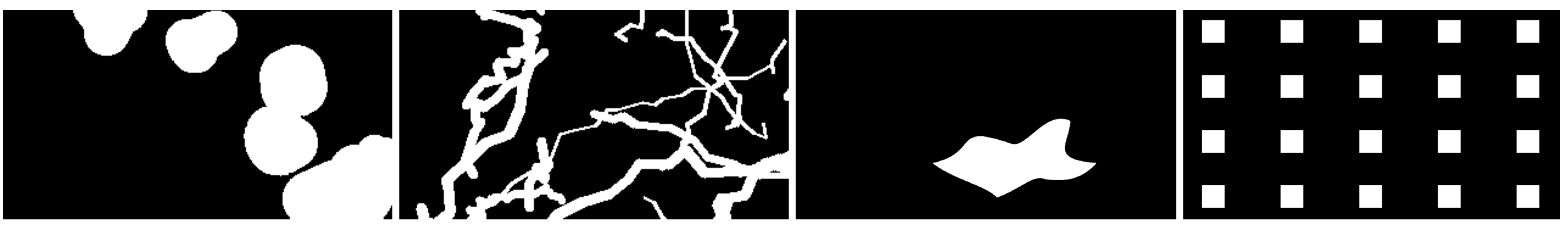}

	\caption{
	Four different types of mask. 
	In FVI \cite{zou2021progressive} dataset, they use (a),(b) as a moving mask and (c) as a stationary mask.
	In our Youtube-VI dataset, (c) and (d) is used as a moving mask and stationary mask respectively.
	}
	\label{fig:masks}
\end{figure}

TSAM \cite{zou2021progressive} publicly provides frame-wise masks for the FVI \cite{yu2019free} dataset.
It consists of 100 scenes, each with 15 frames. 
For each scene, three types of masks are provided, some of which are shown in \fref{fig:masks}.
While the masks cover diverse scenarios, 15 frames in a video are still limited compared to real-world videos.
Also, the problems mentioned above on the mask still exist.

To this end, we design the new Youtube Video Inpainting (Youtube-VI) dataset, which will be publicly available.
100 scenes are selected in order of name from the beginning of Youtube-VOS dataset, and each scene has 50 frames.
We generated two types of masks, which are shown in \fref{fig:masks} and described as follows:

\noindent (1) Moving mask. 
We follow the same rules for generating free-form mask in STTN \cite{zeng2020learning}.
But we use a video segmentation network STCN \cite{cheng2021stcn} so that an object is not fully covered.
With the help of STCN, we randomly generate one moving mask that partially overlaps with objects.
Then we modify the mask again if the scene and mask do not look visually proper. 

\noindent (2) Stationary mask.
This case is usually used to cover the logo/subtitle removal scenarios.
We find that using free-form masks is unsuitable since there may be no motion near the mask.
Instead, we use 5x4 small square masks like FGVC \cite{gao2020flow}.



\subsection{Comparisons}
We quantitatively compare with others on published FVI \cite{zou2021progressive} and our new dataset for video restoration scenarios.
Like \cite{zeng2020learning, zou2021progressive, liu2021fuseformer}, we use DAVIS \cite{caelles20182018}  dataset for qualitative comparison in object removal scenarios.
DAVIS dataset consists of 150 videos in total, in which 90 videos are annotated with the pixel-wise object masks.
In addition, DAVIS-shadow dataset \cite{huang2016temporally} where the mask covers both an object and its shadows are used for more visual comparison.

DFC-Net \cite{xu2019deep} and FGVC \cite{gao2020flow} only take original frames as input to the flow estimator from their published code.
It is acceptable in object removal scenarios, but it is not fair to evaluate quantitative performance where the original frame is equal to ground-truth.
Therefore, we also experiment with corrupted frames as input, and denote them as DFC-Net* and FGVC* respectively.

\begin{table*}[t]
\tiny
\centering
\begin{tabular} 
{@{}P{0.15\linewidth}|P{0.07\linewidth}@{}P{0.07\linewidth}|P{0.07\linewidth}@{}P{0.07\linewidth}|P{0.07\linewidth}@{}P{0.07\linewidth}|P{0.07\linewidth}@{}P{0.07\linewidth}|P{0.07\linewidth}@{}P{0.07\linewidth}|P{0.07\linewidth}}
\multicolumn{1}{l|}{} & \multicolumn{6}{c|}{FVI \cite{zou2021progressive}}  & \multicolumn{4}{c|}{Youtube-VI} & \\ \hline
\multicolumn{1}{l|}{} & \multicolumn{2}{c|}{Stationary Mask} & \multicolumn{2}{c|}{Circle Mask} & \multicolumn{2}{c|}{Curve Mask} & \multicolumn{2}{c|}{Stationary Mask} & \multicolumn{2}{c|}{Moving Mask} \\ \cline{2-11}
& PSNR & SSIM & PSNR & SSIM & PSNR & SSIM &PSNR & SSIM  & PSNR & SSIM  & Time(s) \\ \hline
DFC-Net \cite{xu2019deep}    & 29.20  & 0.9632   & 24.77  & 0.9219 &   28.03  & 0.9436 &   29.03  & 0.9609 &   32.94  & 0.9771 &   95 \\
DFC-Net* \cite{xu2019deep}    & 28.51  & 0.9609   & 24.12  & 0.9141 & 26.60 & 0.9295 & 27.73  & 0.9536  & 32.14  & 0.9747  & 95 \\
OPN \cite{oh2019onion}   & 31.59  & 0.9689  & 26.37  & 0.9264 & 29.79  & 0.9519  & 31.91 & 0.9721  & 33.56 & 0.9767 & 154 \\
CPN \cite{lee2019copy}    & 31.76  & 0.9693  & 26.90  & 0.9315  & 31.09  & \underline{0.9664} & 31.32  & 0.9687  & 34.15  & 0.9783 & 24 \\
STTN \cite{zeng2020learning}    & {32.18}  & 0.9699  & 27.04  & 0.9288 & 30.01 & 0.9515 & 31.96  & 0.9716  & 34.66 & 0.9789 & 17 \\
FGVC  \cite{gao2020flow}  & 30.88  & 0.9690   & 26.14 & 0.9369  & \textbf{31.50}  & \textbf{0.9676}  & 32.25  & 0.9750  & \underline{35.68}  & \underline{0.9841}  &  734 \\
FGVC*  \cite{gao2020flow}  & 30.60  & 0.9682   & 25.55  & 0.9302 & 29.90 & 0.9565 & 31.21  & 0.9701  & 34.05  & 0.9795  & 734 \\
TSAM  \cite{zou2021progressive}  & 31.74  & 0.9695   & 26.66  & 0.9250  & \underline{31.18}  & 0.9578 & -  & -  & -  & -  & - \\
FuseFormer  \cite{liu2021fuseformer}  & \underline{33.19}  & \underline{0.9733}   & \underline{27.85}  & \underline{0.9368}  & 30.85 & 0.9577   & \underline{33.04}  & \underline{0.9755}  & 35.29  & 0.9806 & 18 \\
ECFVI(Ours) & \textbf{33.27}  & \textbf{0.9749}  & \textbf{28.24}  & \textbf{0.9469}  & 31.15  & 0.9638  &  \textbf{33.53}  & \textbf{0.9795} & \textbf{36.80}  & \textbf{0.9860}   & 121 \\
\end{tabular}
\caption{
Quantitative evaluation on FVI and Youtube-VI datasets.
Best values are shown in bold and second best values are underlined.
For PSNR and SSIM, the higher is better. 
Missing entries indicate the method impossible to run at those settings.
}
\label{tab:quantitative}
\end{table*}

\paragraph{Quantitative Results.}
We evaluate video inpainting quality on PSNR and SSIM. 
In \Tref{tab:quantitative}, we achieve the best performance on PSNR and SSIM except for the curve mask on FVI dataset. 
It is worth noting that the curve mask is a particular case in real-world scenarios.
As we assume our error guidance map for common scenarios, the error values in dilated regions can be overlapped, which worsens our results in such thin and overlapped masks.
On the other hand, on the moving and stationary mask, our framework clearly outperforms by a large margin compared to DFC-Net \cite{xu2019deep} and FGVC \cite{gao2020flow}, where the flow-based propagation is used.
Such results show that our overall framework plays critical role in addressing the limitations of existing methods.
More comparisons on other metrics, Video-based Fr\'echet Inception Distance (VFID)~\cite{wang2018video} and optical flow based warping error (EWarp)~\cite{wang2018video}, are shown in the supplementary material.



We measure the execution time on 864x480 resolution for inpainting 50 frames to compare the efficiency.
Since we design our entire components with deep frameworks, ours is $\times{6}$ faster compared to FGVC \cite{gao2020flow} where the optimization-based methods are used.


\begin{figure}[t]
	\centering
	
	\scriptsize
	\begin{tabular} 
        {@{\hskip2pt}p{0.19\linewidth}@{\hskip2pt}p{0.19\linewidth}@{\hskip2pt}p{0.19\linewidth}@{\hskip2pt}p{0.19\linewidth}@{\hskip2pt}p{0.19\linewidth}@{}}
        \centering~Corrupted frame & \centering~STTN \cite{zeng2020learning} & \centering~FGVC \cite{gao2020flow} & \centering~FuseFomer \cite{liu2021fuseformer} & \centering~ECFVI(Ours)
    \end{tabular}
    
	\includegraphics[width=1\linewidth]{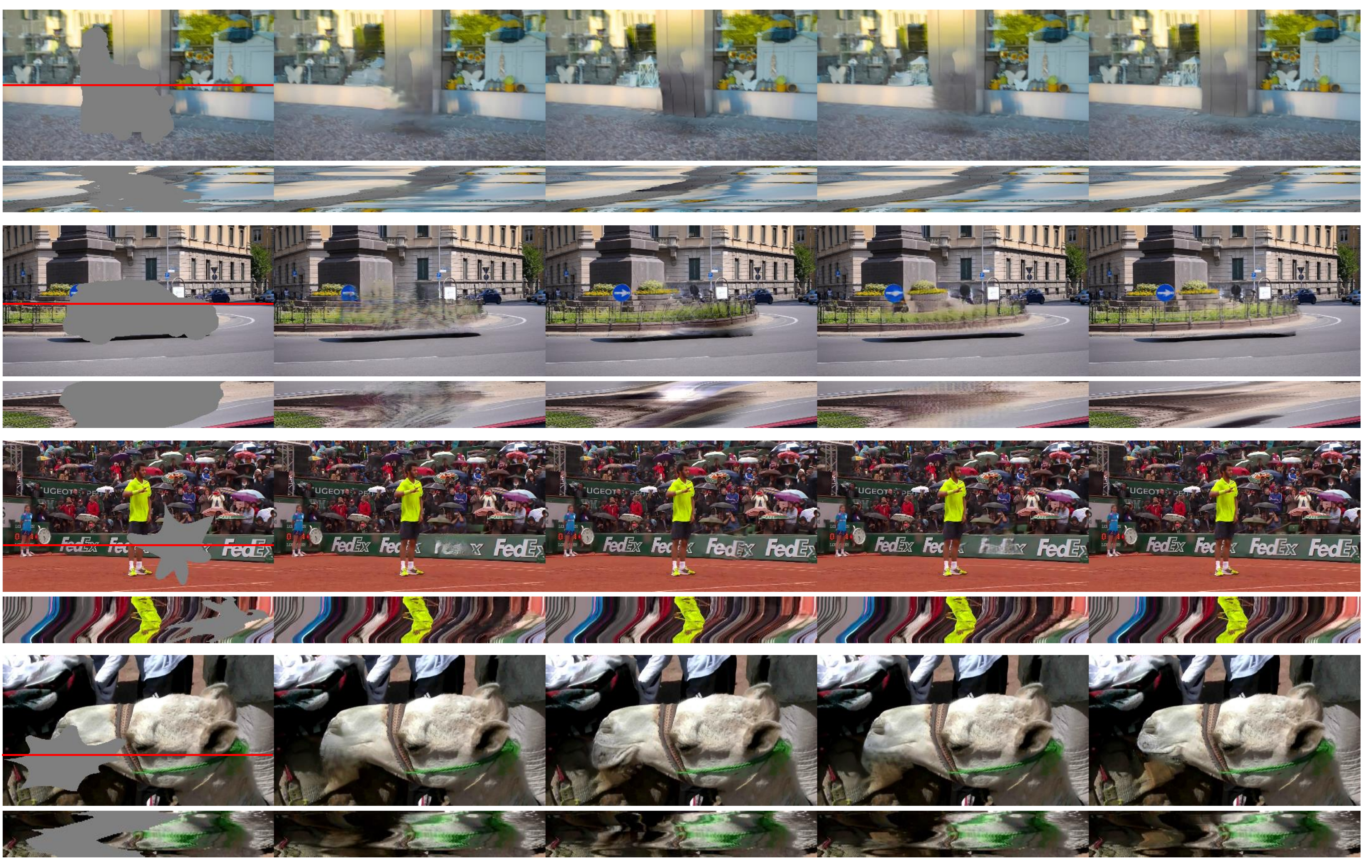}
	\caption{
	Qualitative results compared with \cite{zeng2020learning, gao2020flow, liu2021fuseformer}.  
	The temporal profiles of the red scan line are shown below the results.
	\textbf{Best viewed in zoom.}
	}
	\label{fig:qualitative_davis}
\end{figure}

\begin{figure*}[t]
	\centering
    
	\includegraphics[width=1\linewidth]{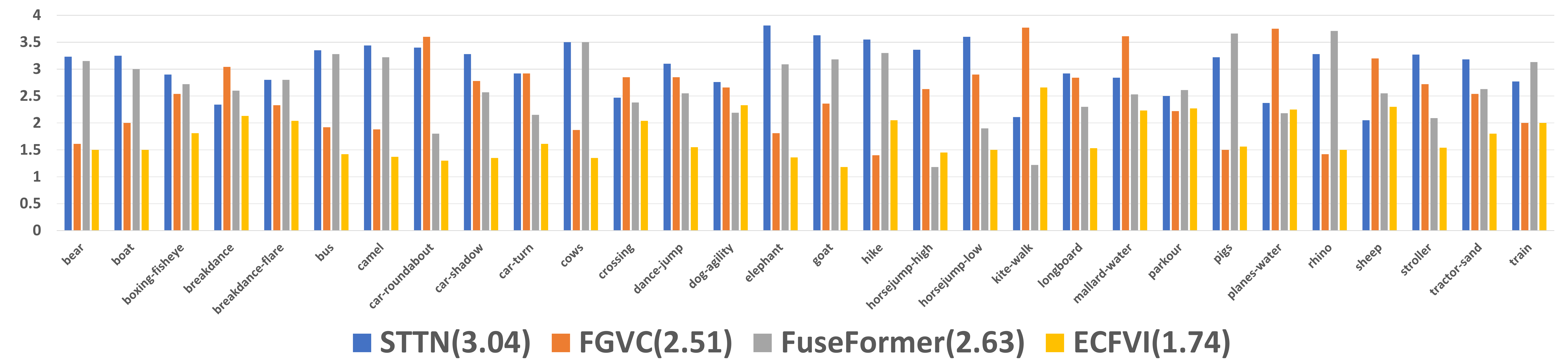}
	\caption{
	A user study on DAVIS \cite{caelles20182018} object removal.
	The lower is better.
	}
	\label{fig:user_study_statics}
\end{figure*}

\paragraph{Qualitative Results.}
In \fref{fig:qualitative_davis}, we show our model's qualitative results compared with other methods, including STTN \cite{zeng2020learning}, FGVC \cite{gao2020flow} and FuseFormer \cite{liu2021fuseformer}.
To visualize the temporal consistency, we show the temporal profile \cite{caballero2017real} of the resulting videos below the completed frames.
Sharp and smooth edges in the temporal profile indicate that the video has much less flickering.
As can be seen, our method shows more visually pleasing results compared to others.

For a more comprehensive comparison, we conducted a user study to subjectively compare our method against others on the DAVIS dataset through Amazon Mechanical Turk.
The experiments were performed on 30 videos that excludes easy (Tennis, Flamingo, etc.) and difficult (India, Drone, etc.) cases where every methods failed. 
The videos like \fref{fig:qualitative_davis} were shown to 20 participants. 
We unlabeled and shuffled the order in all videos, and asked to rank the results from 1 to 4 (1 is the highest preference).
As shown in \fref{fig:user_study_statics}, the results of our model showed higher preference over others, supporting that our approach can produce more visually pleasing outputs.

\begin{table}[t]
    \tiny
    \centering
    
    \begin{subtable}[t]{.52\linewidth}
        \centering
        \caption{Flow completion}
        \label{tab:EPE}
        \begin{tabular}{c|c|c|c}
            \multicolumn{1}{l|}{} & \multicolumn{1}{c|}{Stationary Mask} & \multicolumn{1}{c|}{Moving Mask} &                          \\ \cline{2-4}
            \multicolumn{1}{l|}{} & \multicolumn{1}{c|}{Flow-EPE}             & \multicolumn{1}{c|}{Flow-EPE}            & \multicolumn{1}{c}{Time(s)} \\ \hline
            DFC-Net/* \cite{xu2018youtube}     &  0.32/0.89 &  0.38/0.86 &  43   \\
            FGVC/*  \cite{gao2020flow}     & 0.12/0.57 & 0.28/0.80 & 412 \\
            FF \cite{liu2021fuseformer} + RAFT \cite{teed2020raft}    & 0.49 & 0.80 & 52 \\
            $w/o$ $LTN$     & 0.71 & 0.83 & 24 \\
            $w/o$ Hard     & 0.40 & 0.58 & 52 \\
            Ours           &  0.33 &    0.49  & 52                        
        \end{tabular}
    \end{subtable}%
    \begin{subtable}[t]{.48\linewidth}
        \caption{Video inpainting}
        \label{tab:ablation}
        \centering
        \begin{tabular}
            {@{}P{0.24\linewidth}|P{0.15\linewidth}@{}P{0.15\linewidth}|P{0.15\linewidth}@{}P{0.15\linewidth}}
          & \multicolumn{2}{c|}{Stationary Mask} & \multicolumn{2}{c}{Moving Mask} \\ \cline{2-5} 
          & PSNR       & SSIM         & PSNR      & SSIM      \\ \hline
            FGVC* + Comp & 32.87 & 0.9762 & 35.59    & 0.9823       \\
            GT flow + Comp & 33.64 & 0.9796 & 37.17 & 0.9865       \\
            $w/o$ Comp & 30.05  & 0.9691   & 34.04  & 0.9810      \\   
            $w/o$ Guide  & 30.48   & 0.9692     & 33.81  & 0.9797  \\    
            Ours & 33.53 & 0.9795  & 36.80 & 0.9860     \\
        \end{tabular}
    \end{subtable}%
    \caption{Ablation studies on Youtube-VI dataset. For Flow-EPE, the lower is better.
    FGVC/* means FGVC and FGVC* respectively.}
    \label{ablation study}
\end{table}

\subsection{Ablation Study}
\paragraph{Effectiveness of Flow Completion.}
In \Tref{tab:EPE}, we show the flow end-point-error (EPE) metric to compare our flow completion with others.
We set the estimated flows from RAFT \cite{teed2020raft} as pseudo ground-truth flows.
In the table, FGVC/* means FGVC and FGVC* respectively.
In DFC-Net \cite{xu2019deep} and FGVC \cite{gao2020flow}, the performance is significantly reduced when the corrupted frames are used as input to the flow estimator.
It verifies that the errors in corrupted flows prevent estimating correct flows.
In the same setting where the corrupted frames are used, ours achieves the best performance.

In addition, we conduct a series of ablation studies to analyze the effectiveness of our flow completion module.
First, we sequentially run pretrained FuseFormer \cite{liu2021fuseformer} followed by original RAFT \cite{teed2020raft} (FF + RAFT).
To demonstrate the effectiveness of local temporal network $LTN$, we show the results only using the flow estimator ($w/o$ $LTN$).
The flow estimator is trained with the same setting as ours.
In addition, we train our flow estimator with equal weights in \eref{eq:loss of flow spec} instead of $h$ ($w/o$ Hard).
Such results verify that our jointly trained flow completion module can deal with the errors from the local temporal network $LTN$ and produce more accurate completed flows.

\paragraph{Effectiveness of Compensation.}
In \Tref{tab:ablation}, we show the performance of the error compensation network by replacing our flow completion with FGVC's method (FGVC* + Comp).
Although the flow completion from FGVC* is worse than ours, it outperforms the full procedure of FGVC*.
We also show the pseudo ground-truth flow from RAFT \cite{teed2020raft} (GT flow + Comp).
These two experiments demonstrate our error compensation network is robust to errors from the flow completion. 
Also, we test our model without compensation stage ($w/o$ Comp) and without error guidance map ($w/o$ Guide).
In $w/o$ Guide experiment, we use naive GAN where filled frames with propagation masks and remaining masks are only used as input.
The results show severe performance drops compared to our method.
These two experiments demonstrate that our compensation network using the error guidance map is necessary to handle the errors from the previous stages.


	
    

\section{Conclusion}
In this paper, we have proposed a simple yet effective video inpainting framework, which takes advantage of the flow-based methods while compensating for its shortcomings.
We can propagate valid pixels with our flow completion and error compensation network, showing high-quality completed videos.
Especially with the help of the error guidance map, we can prevent the errors from accumulating and amplifying at the following stages.
We show that our method achieves state-of-the-art performance and demonstrate the benefits of our framework through extensive experiments.
We also provide a new benchmark dataset, which provides comparative analysis into video inpainting methods.

\paragraph{Acknowledgement} This work was supported by Institute of Information \& Communications Technology Planning \& Evaluation (IITP) grant funded by the Korea government (MSIT) (No. 2014-3-00123, Development of High
Performance Visual BigData Discovery Platform for LargeScale Realtime Data Analysis), and No. 2020-0-01361, Artificial Intelligence Graduate School Program (Yonsei University).

%
%





\bibliographystyle{splncs04}
\bibliography{egbib}
\end{document}